# Enhancing VAEs for Collaborative Filtering: Flexible Priors & Gating Mechanisms

Daeryong Kim
Human Centered Computing Lab
Seoul National University
daeryong@snu.ac.kr

Bongwon Suh
Human Centered Computing Lab
Seoul National University
bongwon@snu.ac.kr

## ABSTRACT

Neural network based models for collaborative filtering have started to gain attention recently. One branch of research is based on using deep generative models to model user preferences where variational autoencoders were shown to produce state-of-the-art results. However, there are some potentially problematic characteristics of the current variational autoencoder for CF. The first is the too simplistic prior that VAEs incorporate for learning the latent representations of user preference. The other is the model's inability to learn deeper representations with more than one hidden layer for each network.

Our goal is to incorporate appropriate techniques to mitigate the aforementioned problems of variational autoencoder CF and further improve the recommendation performance. Our work is the first to apply flexible priors to collaborative filtering and show that simple priors (in original VAEs) may be too restrictive to fully model user preferences and setting a more flexible prior gives significant gains. We experiment with the VampPrior, originally proposed for image generation, to examine the effect of flexible priors in CF. We also show that VampPriors coupled with gating mechanisms outperform SOTA results including the Variational Autoencoder for Collaborative Filtering by meaningful margins on 2 popular benchmark datasets (MovieLens & Netflix).

## 1 INTRODUCTION

Today, the immense size and diversity of Web-based services make it nearly impossible for individual users to effectively search and find online content without the help of recommender systems.

There have been various kinds of recent studies incorporating deep learning into recommender systems. We focus on the branch of research using autoencoders and generative models which model latent variables of user preference [15, 23, 30]. Recommendation can be done by using the latent variables of a given user to reconstruct the users' history for recommendation. There has been work using vanilla autoencoders [23], denoising autoencoders [30], and most recently Variational Autoencoders (VAEs) [15] to model user preference for collaborative filtering. To the best of our knowledge, Variational Autoencoders for Collaborative Filtering currently gives state-of-the-art results in the context of collaborative filtering.

However, while many new variations of VAEs are being proposed in the domain of image and audio generation, there has not yet been much research that has yielded further success in the collaborative filtering task for recommender systems.

In this work we aim to overcome some potentially problematic characteristics of VAEs in the task of collaborative filtering and appropriately tailor VAEs to further improve model performance and make high quality recommendations.

Two main motivations led our research. 1) The current prior distribution used in VAEs may be too restrictive for the collaborative filtering task, hindering the models from learning richer latent variables of user preference which is crucial to model performance. 2) Learning from user-item interaction history has its own characteristics and may have more effective architectures to learn deeper latent representations.

We implement hierarchical variational autoencoders with VampPrior (variational mixture of posteriors prior) [25] to learn richer latent representations of user preferences from interaction history. Another variation we adopted is that we used Gated Linear Units (GLUs) [4] to effectively control information flow of our networks by learning when each item or feature contribute to certain units. Coupling the gating mechanism with the aforementioned VampPrior significantly boosted the performance of the variational autoencoding CF framework and outperformed current state-of-the-art collaborative filtering algorithms.

We carried out rigorous experiments on two popular benchmark datasets: MovieLens-20M and Netflix. Our proposed method was compared to baseline models including state-of-the-art matrix factorization and autoencoder based methods and showed significant improvements in NDCG and recall.

The key contributions of our work are as follows:

- Our work is the first to address the restrictive prior problem for the VAE-CF framework and shows that relaxing the prior to a more flexible distribution yields better recommendation performance.
- We show introducing gating mechanisms are also very helpful for autoencoder based CF in learning deeper and more sophisticated representations of interaction history.
- Our proposed model using hierarchical VAEs with VampPrior and Gated Linear Units gives new State-Of-The-Art results on standard benchmark datasets in the task of collaborative filtering.

## 2 RELATED WORK

There have been various studies incorporating deep learning into collaborative filtering recommender systems. Research extending the traditional matrix factorization framework to non-linear matrix factorization using neural networks [9], session-based recommenddation using recurrent neural networks (RNNs) [10, 20, 29], recommendation with autoencoders and generative models [15, 23, 28, 30], and many others including hybrid methods using extraction of high-level content features through deep learning [26, 27].

The autoencoder based recommendation algorithm was first proposed as AutoRec [23]. It is the algorithm of using vanilla autoencoders for collaborative filtering and showed to give superior results compared to linear MF methods. Further research was made using denoising autoencoders to present CDAE [30].

The most recent advancement of autoencoder based CF was made by using Variational Autoencoders for Collaborative Filtering [15]. The Variational Autoencoder (VAE) is a probabilistic generative model in the form of autoencoders which models the data distribution P(X) using amortized variational inference. In VAE-CF [15] the latent variables are stochastic, and their probability distributions are learned for each datapoint. Additionally modeling per-data-point variation led to more robust representations and yielded SOTA recommendation performance beating other autoencoder and neural network based methods such as CDAE [30] and Neural Collaborative Filtering (NCF) [9].

On the other hand, in the domain of computer vision there have been further advances for VAEs proposing new models with more flexible priors to enrich the generative capabilities. A Dirichlet process prior with a stick-breaking process was proposed in [18], and in [8] a nested Chinese Restaurant Process was used. Also, a Gaussian mixture prior was used for [6]. The VampPrior [25] was proposed recently, it consists of a mixture distribution of variational posteriors on pseudo-inputs substituting the original standard normal prior to a very flexible multimodal distribution. The VampPrior showed impressive results for image generation and was a major motivation of our work.

## 3 PRELIMINARIES

Our work is an extension of the VAEs for CF [15] framework incorporating appropriate ideas to further enhance the recommendation performance in collaborative filtering. In this section we describe our problem formulation and the original VAE-CF framework.

### 3.1 Problem Formulation

We attempt to model user preferences based on a given users' interaction history of the item set. We shall use the following shared notations throughout the paper. We will use $u \in \{1, \ldots, N\}$ to index users and $i \in \{1, \ldots, M\}$ to index items. We consider implicit feedback with binary input: the dataset $\mathbf{X} = \{x_1, \ldots, x_N\}$ with each $x_u \in \mathbb{I}^M$ the interaction history of user $u$. And $z_u \in \mathbb{R}^D$ the latent variable of user preference for user $u$.

### 3.2 VAE for Collaborative Filtering

The baseline model of our research is the Multi-VAE in [15]. The generative process of the model is as follows. For every user $u$ a latent variable $z_u \in \mathbb{R}^D$ is sampled from the standard normal prior distribution. The latent representation is then transformed through a neural network generative model to produce the probability distribution over the user's item consumption history $x^u$, a bag-of-words vector indicating whether the user has consumed each item, assuming a multinomial distribution:

$$z_u \sim N(0, I_D), \quad \pi(z_u) \propto \exp\{f_\theta(z_u)\}$$
$$x_u \sim Multi(N_u, \pi(z_u)) \qquad (1)$$

Once the generative process is configured, it follows the typical Variational Autoencoder [14] framework and attempts to maximize the marginal data likelihood $P(X) = \int p(X|z)p(z)\,dz$. Since a Neural Network is used for the non-linear mapping $f_\theta(\cdot)$, $P(X)$ becomes intractable and the optimization becomes difficult.

The problem is solved by using amortized variational inference and optimizing per-datapoint the following Evidence Lower Bound (ELBO) [14]:

$$\log p(x_u; \theta) \geq \mathbb{E}_{q_\phi(z_u|x_u)}[\log p_\theta(x_u|z_u)]$$
$$- \mathrm{KL}\left(q_\phi(z_u|x_u) || p(z_u)\right)$$
$$\equiv \mathcal{L}(x_u; \theta, \phi) \qquad (2)$$

with $p_\theta(x|z)$ a *generative model (decoder)* which is a neural network parameterized by $\theta$, a *prior distribution* of latent variables $p_\lambda(z)$, and an approximation to the unknown posterior $p(z|x)$ with a *recognition model (encoder)* $q_\phi(z|x)$ also with neural networks. The Multi-VAE for CF [15] additionally introduces a parameter $\beta \in [0,1]$ to scale the KL term similar to $\beta$-VAE [3].

## 4 ENHANCING VAES FOR CF

Variational Autoencoders have been extensively researched in the fields such as image generation and new advances have been proposed since its first appearance [14]. One line of research analyzes the prior distribution of VAEs, suggesting that regular standard Gaussian priors can restrict the modeling performance [11, 16]. However, the restrictive prior problem has not been researched in the field of recommender systems and our work is the first to apply flexible priors to variational autoencoders for collaborative filtering.

### 4.1 Flexible Priors for Modeling User Preference

As in [11, 16, 25], the ELBO objective can be further analyzed to be rewritten as the following:

$$\mathcal{L}(\theta, \phi, \lambda) = \mathbb{E}_{x \sim q(x)}\left[\mathbb{E}_{q_\phi(z|x)}[\log p_\theta(x|z)]\right]$$

$$+\mathbb{E}_{x \sim q(x)}\left[\mathbb{H}[q_\phi(z|x)]\right]$$
$$-\mathbb{E}_{z \sim q(z)}[-\log p_\lambda(z)] \quad (3)$$

The first term is the negative reconstruction error while the second term is the expected entropy of the variational posterior, and the last component is the cross-entropy between the aggregated posterior $q(z) = \frac{1}{N}\sum_{u=1}^{N} q_\phi(z|x_u)$ and the prior.

We can see that the cross-entropy term pulls the distribution of the latent variables towards the prior, which is in the case of regular VAEs a standard Gaussian distribution chosen in advance. This can result in an unintended strong regularization effect due to the simple unimodal nature of the standard Gaussian distribution.

While the encoder tries to shape the aggregated posterior to match the prior distribution, there is no guarantee that a simple unimodal distribution will be a good match. Since modeling human preference is a complicated issue, we considered it plausible to relax the restrictive prior to investigate possible increase of recommendation quality in the context of collaborative filtering.

**VampPrior**. We experiment with a recently proposed flexible prior called the VampPrior (variational mixture of posteriors prior) [25]. Revisiting equation (3), we can see that only the cross-entropy term is associated with the prior $p_\lambda(z)$. If we find the optimal prior maximizing the ELBO by solving the Lagrange function it simply gives us the aggregated posterior $p_\lambda^*(z) = \frac{1}{N}\sum_{u=1}^{N} q_\phi(z|x_u)$. VampPrior [25] is an approximation to this optimal prior using a mixture distribution of variational posteriors conditioned on K learnable pseudo-inputs:

$$p_\lambda(z) = \frac{1}{K}\sum_{k=1}^{K} q_\phi(z|u_k) \quad (4)$$

where $K(\ll N)$ is the number of M-dimensional pseudo-inputs $u_k$. The pseudo-inputs are learned through backpropagation and can be thought of hyperparameters of the prior.

**Hierarchical Stochastic Units**. We also adopt hierarchical stochastic units to learn even richer latent representations as in the original work of VampPriors [25]. Hierarchical VAEs have been explored in different literatures [12, 25] but have not been explored for collaborative filtering.

The original stochastic latent variable $z$ is replaced by a stacked hierarchical structure of $z_1$ and $z_2$. The full Hierarchical VampPrior VAE model is given as the follows. The variational part:

$$q_\phi(z_1|x, z_2) \, q_\psi(z_2|x) \quad (5)$$

and the generative part:

$$p_\theta(x|z_1, z_2) \, p_\lambda(z_1|z_2) \, p(z_2) \quad (6)$$

where $p(z_2)$ is given as a VampPrior $p(z_2) = \frac{1}{K}\sum_{k=1}^{K} q_\phi(z_2|u_k)$ and other conditional distributions are each modeled by neural networks.

## 4.2 Gating Mechanism

Preceding research using autoencoders for collaborative filtering make use of relatively shallow networks. Models in [23, 30] use encoder networks with no hidden layers. The encoder for Multi-VAE [15] use 1 hidden layer and does not achieve additional performance gain by adding more layers. We anticipate two possible reasons for this; (1) the nature of the data, where we have to extract preference from sparse consumption history and (2) the relatively *easily deepening* autoencoder structure due to the encoder and decoder.

**Gated Linear Units**. As the structure of Neural Networks get deeper and deeper, non-recurrent neural nets also have the problem of being unable to properly propagate information from the bottom layer to the top. We experiment with a non-recurrent gating mechanism proposed in Gated CNNs [4] which was suggested to help information propagation in deeper networks:

$$h_l(X) = (X * W + b) \otimes \sigma(X * V + c) \quad (7)$$

$\otimes$ is the element-wise product with $X$ the input of the layer and $W, V, b, c$ learned parameters, and $\sigma$ the sigmoid function. As we can see from the formula, the gates attend to the past layer and react depending on the current input. This can also be interpreted as potentially increasing the network's modeling capacity to allow higher level interactions.

## 5 EXPERIMENTS

Experiments were conducted to evaluate the effect of flexible priors, hierarchical stochastic units and gating mechanisms in the context of collaborative filtering. Our proposed models are compared to other state-of-the-art collaborative filtering models. The source code is available on GitHub (http://github.com/psywaves/EVCF).

### 5.1 Setup

**Datasets**. The Experiments were made on the MovieLens-20M and Netflix Prize dataset. Since we consider implicit feedback, we binarize both datasets by keeping only ratings of four or higher. Also, for both datasets we keep only users who have watched at least five movies.

**Metrics**. We evaluate performance based on two ranking-based metrics: Recall@K and truncated normalized discounted cumulative gain (NDCG@K). Recall@K considers all items ranked within the first K to be equally important, while NDCG@K uses a monotonically increasing discount to emphasize the importance of higher ranks versus lower ones [15].

**Experimental settings**. Models are evaluated under the strong generalization setting following [15, 17]. All users are split into training/validation/test sets. The models are trained using the entire click history of the training set. During evaluation, we sample 80% of the click history from each user in the validation (or test) dataset as the "fold-in" set to calculate the necessary user-level representations and predict the remaining 20% of the click history.

### 5.2 Models

We use popular matrix factorization and state-of-the-art autoencoder models as baselines for comparison. **WMF** [13], **SLIM** [19], **CDAE** [30] and **Multi-VAE** [15] are chosen as baselines.

Our models to evaluate the effect of flexible priors, HVAE and gating are the following.

**Vamp**: Variational autoencoder with a VampPrior [25] as the prior distribution instead of the original standard normal prior. We can compare with Multi-VAE [15] to evaluate the effect of using flexible priors.

**H + Vamp**: Hierarchical VAE [12, 25] with the VampPrior, the difference to the *Vamp* model is that it has hierarchical stochastic units to model the latent representation.

|  | **MovieLens 20M** | | | **Netflix** | | |
| --- | --- | --- | --- | --- | --- | --- |
| Models | NDCG@100 | Recall@50 | Recall@20 | NDCG@100 | Recall@50 | Recall@20 |
| WMF [13] [†] | 0.386 | 0.498 | 0.360 | 0.351 | 0.404 | 0.316 |
| SLIM [19] [†] | 0.401 | 0.495 | 0.370 | 0.379 | 0.428 | 0.347 |
| CDAE [30] [†] | 0.418 | 0.523 | 0.391 | 0.376 | 0.428 | 0.343 |
| Mult-VAE [15] | 0.42700 | 0.53524 | 0.39569 | 0.38711 | 0.44427 | 0.35255 |
| Vamp | 0.43433 | 0.53933 | 0.40310 | 0.39589 | 0.44907 | 0.36327 |
| H+Vamp | 0.43684 | 0.53974 | 0.40524 | 0.40242 | 0.45605 | 0.37090 |
| Mult-VAE (Gated) | 0.43515 | 0.54498 | 0.40558 | 0.39241 | 0.44958 | 0.35953 |
| **H+Vamp (Gated)** | **0.44522** | **0.55109** | **0.41308** | **0.40861** | **0.46252** | **0.37678** |

**Table 1: Results for MovieLens 20M and Netflix dataset. Standard errors are around 0.002 for ML-20M and 0.001 for Netflix.**
[†]**Results are taken from [15], note that our datasets, metrics and experimental settings are consistent with [15].**

**H + Vamp (Gated)**: Our final model, additional gating mechanisms are applied to the *H + Vamp* above. Gated Linear Units [4] are used for all hidden units in the network.

**Multi-VAE (Gated)**: The Multi-VAE [15] model with gating mechanisms. This model was additionally studied for comparison.

All models are fully tuned by choosing hyperparameters with grid search on possible candidate values[1]. The number of components **K** for the VampPrior was set to 1000. Also, while it was suggested in [15] that multinomial likelihoods perform better for CF than binary cross-entropy, we found it was not always the case and used the better of the two for each model/dataset.

Results of WMF [13], SLIM [19] and CDAE [30] were taken from [15]. Note that our datasets and setup are consistent with [15] for fair comparison.

## 5.3 Results

In this session, we first compare the performance results of our proposed models with the various baselines. We then further examine the effect of gating mechanisms by comparing performance of gated and ungated models of increasing depth.

**Model performance**. Quantitative results comparing performance are presented in Table 1. Multi-VAE [15] is the strongest baseline while Vamp, H+Vamp, H+Vamp (Gated) shows sequentially improving performance. *Vamp* shows significantly better performance compared to *Multi-VAE*, indicating the benefit of changing the restrictive standard normal prior to a flexible VampPrior. Our final model *H + Vamp (Gated)* shows the best performance and significantly outperforms the strongest baseline *Multi-VAE* [15] for both datasets on all metrics. The final model shows up to 6.87% relative increase in recall@20 for the Netflix dataset producing new state-of-the-art results.

**Effect of gating**. We also conducted experiments to further study the effect of using gates. We present the results in ndcg@100 for the Netflix dataset in Table 2. In this experiment the number of hidden units in each layer is fixed to 600[2]. A two layer model means that there are two hidden layers in each of the encoder and decoder.

We can see in Table 2 that for models with no gates, increasing the depth does not bring performance gain while for gated models it does. This can be interpreted that gating does help the network to propagate information through deeper models. However, we can also see large performance gains in simply adding the gates without additional layers. This tells us that the higher-level interactions the self-attentive gates allow are also very helpful themselves for modeling user preferences. One may point out that the gated model has more parameters, but note that ungated models cannot achieve similar performance by merely adding more units.

| Netflix (NDCG@100) | No-Gate | Gated |
| --- | --- | --- |
| Mult-VAE (1 Layer) | **0.38711** | 0.39229 |
| Mult-VAE (2 Layer) | 0.38359 | **0.39241** |
| Vamp (1 Layer) | **0.39589** | 0.40169 |
| Vamp (2 Layer) | 0.39346 | **0.40277** |
| H + Vamp (1 Layer) | **0.40242** | 0.40728 |
| H + Vamp (2 Layer) | 0.37970 | **0.40861** |

**Table 2: Comparison of performance between Gated and Un-Gated for models of different depth[3]. The model with better performance (1 Layer vs 2 Layers) is marked in bold.**

## 6 CONCLUSION

In this paper, we extend the VAE for collaborative filtering to adopt flexible priors and gating mechanisms. We show empirically that standard Gaussian priors may limit the model capacity and introducing a more flexible prior can learn better representations of the user preference. We also show that gating mechanisms are suitable for user-item interaction data. Gates provide valuable modeling capacity as well as helping information propagate through deeper networks.

Our final model incorporating Hierarchical VampPrior VAEs with GLUs produces new state-of-the-art results in the collaborative filtering literature.

## Acknowledgements
Part of this work was done during internship at Kakao corporation, Republic of Korea.

---

[1] Hyperparameters were selected using the validation set.
[2] All other hyperparameters except the number of layers were fixed as well.
[3] There was no additional performance gain for adding more hidden layers than two.